\documentclass[]{fairmeta}

\usepackage{hyperref}
\usepackage{url}
\usepackage{booktabs}
\usepackage{multirow}
\usepackage{amsmath}
\usepackage{amssymb}
\usepackage{amsthm}
\usepackage{algorithm}
\usepackage[noend]{algpseudocode}
\usepackage{graphicx}
\usepackage{wrapfig}
\usepackage{placeins}
\usepackage[most]{tcolorbox}
\usepackage[table]{xcolor}

\usepackage{enumitem}

\title{Bidirectional Across Items,
  Sequential Within IDs: Hierarchy-Preserving Masked Discrete Diffusion for
  Generative Recommendation}

\author[1,2,*]{Katherine Tieu}
\author[1]{Dongqi Fu} 
\author[1]{Si Zhang}
\author[1]{Hanqing Zeng}
\author[1]{Chenglin Wei}
\author[1]{Yinglong Xia}
\author[1]{Hong Li}
\author[2]{Jingrui He}

\affiliation[1]{Meta}
\affiliation[2]{University of Illinois Urbana-Champaign}

\contribution[*]{Work done at Meta}

\abstract{%
Generative recommendation represents items as hierarchical semantic IDs (SIDs)
and models next-item prediction as sequence generation. Autoregressive language
models respect the coarse-to-fine order imposed by residual quantization, but
apply the same causal factorization across items in the entire interaction sequence.
Masked discrete diffusion enables bidirectional learning across items; however,
independently masking SID's components can expose a fine residual code while
hiding the coarse prefix on which that code depends. We introduce
\textbf{H}ierarchy-\textbf{P}reserving \textbf{D}iffusion (\textbf{HPD}), a
masked discrete diffusion language modeling formulation that separates
these two dependency levels: (1) \textit{Across SIDs}, full self-attention captures relations
throughout the interaction sequence. (2) \textit{Within each SID}, prefix-independent
masking restricts every corrupted sequence to an observed prefix followed by a masked
suffix. HPD combines this structured corruption with a proposed training
pipeline that first optimizes only the newly added SID-token embeddings and
output weights while the backbone is fixed and then jointly optimizes all
model parameters. A catalog-constrained beam search decodes the target SID
from coarse to fine.
HPD requires no task-specific architectural modification.
Across three domains, Industrial and Scientific, Office Products, and Baby
Products, we compare the
Qwen3-8B/LLaDA-8B-Base and Qwen2.5-7B/Dream-7B-Base autoregressive/diffusion
large language model (AR-LLM/dLLM) pairs. At the final checkpoint after
training concludes, HPD achieves higher Hit Rate and NDCG than its matched
AR-LLM counterpart at every evaluated cutoff.
In the 8B LLaDA comparison, HPD obtains the best or tied-best score across all
18 reported metric--domain combinations. Comparisons with standard MDLM and
DiffGRM evaluate HPD against alternative masked discrete diffusion objectives.
An ablation shows that the proposed HPD pipeline outperforms the SID-token-only
variant on every metric for Dream-7B-Base on Industrial and Scientific and
LLaDA-8B-Base on Baby Products.
}

\date{\today}

\begin{document}

\maketitle

\section{Introduction}
\label{sec:introduction}

Next-item recommendation is central to Web platforms, where recommender systems
organize access to products and content. Within the Search, Recommendation, and
Retrieval-Augmented AI track, we study how models can capture both the
collaborative structure of a user's interaction history and the semantic
relationships among items. Traditional sequential recommenders encode
interaction histories with
recurrent networks or causal self-attention and score items from a fixed
catalog~\citep{hidasi2016session,kang2018self}. Generative recommendation with
semantic IDs (SIDs) instead provides a compact way to combine collaborative and
semantic signals: a quantizer maps each item to a short sequence of discrete
codes, and a sequence model generates the SID of the next
item~\citep{rajput2023recommender}. In the widely
used residual-quantization construction, the first SID component represents a
coarse region of the item-embedding space, while each later component encodes
the residual left by the preceding components. The resulting representation is
therefore not merely a set of interchangeable tokens; it is a hierarchical,
coarse-to-fine description of an item.

Building on this formulation, OneRec unifies retrieval and ranking within an
end-to-end generative system, while QARM aligns multimodal item information
with quantized representations~\citep{deng2025onerec,luo2025qarm}. More recent
work scales the same autoregressive paradigm with general-purpose LLM
backbones. MiniOneRec provides an open pipeline for training Qwen models from
0.5B to 7B parameters with supervised and reinforcement-learning objectives, and
OpenOneRec develops 1.7B- and 8B-parameter recommendation foundation models
through co-pretraining and post-training~\citep{kong2025minionerec,zhou2025openonerec}.
These efforts show how large autoregressive backbones can transfer language
model capacity to recommendation, while retaining a causal token
factorization.

Most SID-based recommenders flatten the interaction sequence and train an
autoregressive language model (AR-LLM) to predict one token at a time. This
factorization is well matched to the dependency within an SID because a finer
component is generated after its coarser prefix. However, it applies the same
left-to-right restriction between different items, even though that order
serves a different role. Within an SID, the order is imposed by the
representation itself; across SIDs, a non-causal objective can use the entire
interaction sequence to learn global relations among items.

Masked discrete diffusion provides such a non-causal alternative. Early work
on discrete diffusion defined denoising processes over categorical state
spaces~\citep{austin2021structured}, and score-entropy discrete diffusion
subsequently formulated training through ratios of the data
distribution~\citep{lou2024discrete}. For language modeling, an absorbing-state
process corrupts discrete tokens by replacing them with a mask. Masked discrete
diffusion language modeling (MDLM) derives a simplified likelihood-based
objective for this process and reconstructs the masked tokens with full
self-attention~\citep{sahoo2024simple}. Unlike continuous diffusion over
embeddings, this formulation operates directly on discrete vocabulary symbols.

Scaling the MDLM family has produced diffusion large language models (dLLMs)
such as LLaDA and Dream~\citep{nie2025llada,ye2025dream}. Subsequent work has
extended these models through variance-reduced preference optimization in
LLaDA~1.5~\citep{zhu2025llada15} and improved from-scratch pretraining in
iLLaDA~\citep{nie2026illada}. Throughout this paper, \emph{MDLM} refers to the
probabilistic family and training objective, whereas \emph{dLLM} refers to a
large pretrained language-model instantiation. We use LLaDA-8B-Base and
Dream-7B-Base as the dLLM backbones in our empirical study.

Applying standard masked discrete diffusion directly to hierarchical SIDs
creates a different mismatch. Independent token masking can hide a coarse
component while leaving one of its finer residual components visible. The
model is then trained to recover a parent code from information that, under
the SID construction, is defined only after that parent is known. Recent
masked discrete diffusion recommenders demonstrate the value of
non-autoregressive generation, but encode recommendation structure in different
ways. MaskGR permits independently masked SID tokens and flexible generation
orders~\citep{shah2025maskgr}. MindRec uses a dLLM to generate salient tokens
before refining a category hierarchy and item representation from coarse to
fine~\citep{gao2025mindrec}. DiffGRM instead replaces residual quantization
with parallel semantic encoding to remove sequential dependence among SID
components~\citep{liu2026diffgrm}. These approaches leave open a complementary
question: how can a dLLM retain the hierarchy already present within residual
SIDs while learning bidirectional relations across items?

We address this question with \emph{Hierarchy-Preserving Diffusion} (HPD), an
MDLM formulation that separates inter-SID and intra-SID dependencies. Across
SIDs, the model uses full self-attention and can condition reconstruction on
observed items at either side of a masked position. Within each SID, HPD's
\emph{prefix-independent masking} process restricts every corrupted SID to an
observed prefix followed by a masked suffix. It first samples raw Bernoulli
mask indicators and then applies a cumulative-maximum filter, so masking any
component also masks all finer components. This structured corruption changes
only the training distribution and can therefore be applied to an existing
dLLM without modifying its architecture. The proposed training pipeline first
optimizes only the new SID-token embeddings and output weights while the
backbone is fixed and then jointly optimizes all model parameters. At inference
time, a catalog-constrained beam search unmasks the target
SID from coarse to fine, matching the dependency enforced during training and
ensuring that every completed SID identifies a catalog item.

We evaluate HPD on three Amazon recommendation domains with two
approximately size-matched AR-LLM/dLLM pairs: Qwen3-8B versus LLaDA-8B-Base
and Qwen2.5-7B versus Dream-7B-Base. At the final checkpoint after training
concludes, HPD achieves higher HR and NDCG than its matched AR-LLM counterpart
at every evaluated cutoff in all three domains. Comparisons with standard
MDLM training and DiffGRM evaluate HPD against alternative masked discrete
diffusion objectives. In the 8B LLaDA comparison, HPD obtains the best or
tied-best score across all 18 reported metric--domain combinations. The
proposed HPD pipeline also outperforms the SID-token-only variant on every
metric for Dream-7B-Base on Industrial and LLaDA-8B-Base on Baby.

Our contributions are as follows:
\begin{itemize}
    \item We distinguish the bidirectional relations between item SIDs from
    the coarse-to-fine dependency within a residual SID, and identify why
    independent token masking can violate this coarse-to-fine dependency
    (Section~\ref{sec:two-level-dependency}).
    \item We introduce HPD with prefix-independent masking
    (Section~\ref{sec:prefix-independent-masking}), a single training run in
    which the set of trainable parameters expands from the new SID-token
    parameters to all model parameters (Section~\ref{sec:hpd-training}), and
    hierarchy-preserving catalog-constrained inference
    (Section~\ref{sec:hpd-inference}). Together, these components apply masked
    discrete diffusion modeling to residual SIDs without changing the backbone
    architecture.
    \item We provide a controlled empirical comparison across two AR-LLM/dLLM
    backbone pairs and three recommendation domains, showing consistent gains
    over matched autoregressive training. HPD obtains the best or tied-best
    score across all 18 reported metric--domain combinations in the 8B comparison.
    We further show that the proposed HPD pipeline outperforms the
    SID-token-only variant on every metric for Dream-7B-Base on Industrial and
    LLaDA-8B-Base on Baby
    (Sections~\ref{sec:main-results} and~\ref{sec:ablation-study}).
\end{itemize}

\section{Preliminaries}
\label{sec:preliminaries}

We first formulate next-item recommendation over hierarchical semantic IDs
(SIDs), and then review the autoregressive and masked discrete diffusion
language-modeling objectives. These definitions establish a common sequence
representation for both objectives and make explicit the masking assumption
modified by HPD.

\subsection{Next-Item Recommendation with Hierarchical SIDs}
\label{sec:problem-formulation}

Let $\mathcal{I}$ denote the item catalog. Each item $i\in\mathcal{I}$ has a
textual description $d_i$, which a fixed text encoder $g_{\mathrm{enc}}$ maps
to an embedding $\mathbf{h}_i=g_{\mathrm{enc}}(d_i)\in\mathbb{R}^d$, where
$d$ is the embedding dimension. We discretize these embeddings with
residual-quantized K-means (RQ-KMeans), which assigns a sequence of
coarse-to-fine codes to every
item~\citep{chen2010approximate,luo2025qarm}. Specifically, let
$\mathcal{C}^{(r)}=\{\mathbf{c}^{(r)}_1,\ldots,
\mathbf{c}^{(r)}_{K_r}\}\subset\mathbb{R}^d$ be the codebook at hierarchy
level $r\in\{1,\ldots,m\}$, where $K_r$ is the number of centroids at that
level and $m$ is the number of hierarchy levels. Starting from
$\mathbf{r}_i^{(1)}=\mathbf{h}_i$, RQ-KMeans recursively assigns
\begin{align}
    s_i^{(r)}
    &= \arg\min_{a\in\{1,\ldots,K_r\}}
       \left\|\mathbf{r}_i^{(r)}-\mathbf{c}^{(r)}_a\right\|_2^2,
       \label{eq:rq-assignment}\\
    \mathbf{r}_i^{(r+1)}
    &= \mathbf{r}_i^{(r)}-\mathbf{c}^{(r)}_{s_i^{(r)}}.
       \label{eq:rq-residual}
\end{align}
Here, $\|\cdot\|_2$ is the Euclidean norm,
$\mathbf{r}_i^{(r)}\in\mathbb{R}^d$ is the residual entering level $r$, and
$s_i^{(r)}$ is the selected centroid index, represented in the language model
by a level-specific SID token. The resulting item representation is
\begin{equation}
    \operatorname{SID}(i)
    = \mathbf{s}_i
    = \left(s_i^{(1)},\ldots,s_i^{(m)}\right).
    \label{eq:sid-definition}
\end{equation}
Because each level quantizes the residual left by earlier levels, the component
order in $\mathbf{s}_i$ is sequential: early components encode coarse
information, while later components provide progressively finer refinements.
We add the level-specific SID tokens to the language model vocabulary. In our
setting, $m=3$, so every item is represented by three SID tokens.

For a user with $n$ observed interactions followed by a target interaction,
let $(i_1,\ldots,i_n,i_{n+1})\in\mathcal{I}^{n+1}$ be the chronologically
ordered item sequence. We represent the observed history and next-item target
as
\begin{equation}
    Q=[\mathbf{s}_1,\ldots,\mathbf{s}_n],
    \qquad
    Y=\mathbf{s}_{n+1}
      =\left(s_{n+1}^{(1)},\ldots,s_{n+1}^{(m)}\right),
    \label{eq:recommendation-pair}
\end{equation}
where $\mathbf{s}_j=\operatorname{SID}(i_j)$ for
$j\in\{1,\ldots,n+1\}$. Let $\mathcal{D}$ denote the empirical distribution
of training pairs $(Q,Y)$. Both language-modeling objectives operate on the
flattened concatenation
\begin{equation}
    \mathbf{x}
    = \operatorname{flat}(Q\oplus Y)
    = (x^1,\ldots,x^L),
    \qquad L=m(n+1),
    \label{eq:flattened-sid-sequence}
\end{equation}
where $\oplus$ denotes concatenation, $\operatorname{flat}$ removes the SID
tuple boundaries while preserving their order, and $x^k$ is the token at
flattened position $k\in\{1,\ldots,L\}$.

\subsection{Autoregressive Language Modeling}
\label{sec:ar-preliminaries}

An autoregressive language model (AR-LLM) imposes a left-to-right factorization
on the flattened sequence. Let $\pi_\theta$ denote the token distribution
parameterized by $\theta$, and define
$\mathbf{x}^{<k}=(x^1,\ldots,x^{k-1})$ as the prefix preceding position $k$.
The autoregressive training objective is
\begin{equation}
    \mathcal{L}_{\mathrm{AR}}(\theta)
    = -\mathbb{E}_{(Q,Y)\sim\mathcal{D}}
      \left[
        \sum_{k=1}^{L}
        \log\pi_\theta\!\left(x^k\mid\mathbf{x}^{<k}\right)
      \right].
    \label{eq:ar-objective}
\end{equation}
Equation~\eqref{eq:ar-objective} trains every token to depend only on its
preceding tokens. Applied to SIDs, this factorization enforces sequential order
both between items and among the $m$ components of each item.

\subsection{Masked Discrete Diffusion Language Modeling}
\label{sec:mdlm-preliminaries}

A masked discrete diffusion language model (MDLM) instead learns to reconstruct
tokens removed by a stochastic masking process~\citep{sahoo2024simple}.
Given $\mathbf{x}$, we sample a masking rate $t\sim\mathcal{U}[0,1]$ from the
continuous uniform distribution on $[0,1]$ and independently replace each
discrete token $x^k$ with a special mask token $\mathrm{M}$ with probability
$t$. This absorbing-state corruption operates on vocabulary symbols; it does
not add continuous noise to token embeddings. The resulting corrupted sequence
is $\mathbf{z}_t=(z_t^1,\ldots,z_t^L)$, drawn from
\begin{equation}
    q_t(\mathbf{z}_t\mid\mathbf{x})
    = \prod_{k=1}^{L}q_t(z_t^k\mid x^k),
    \qquad
    q_t(z_t^k\mid x^k)
    =
    \begin{cases}
        t, & z_t^k=\mathrm{M},\\
        1-t, & z_t^k=x^k,\\
        0, & \text{otherwise}.
    \end{cases}
    \label{eq:mdlm-forward-process}
\end{equation}
The MDLM uses full, non-causal attention and minimizes cross-entropy only at
masked positions:
\begin{equation}
    \mathcal{L}_{\mathrm{MDLM}}(\theta)
    = -\mathbb{E}_{\substack{(Q,Y)\sim\mathcal{D},\;t\sim\mathcal{U}[0,1],\\
        \mathbf{z}_t\sim q_t(\cdot\mid\mathbf{x})}}
      \left[
        \frac{1}{t}\sum_{k=1}^{L}
        \mathbb{I}[z_t^k=\mathrm{M}]
        \log\pi_\theta\!\left(x^k\mid\mathbf{z}_t\right)
      \right],
    \label{eq:mdlm-objective}
\end{equation}
where $\mathbb{I}[\cdot]$ is the indicator function. The factor $1/t$ weights
the reconstruction loss by the sampled corruption level. Unlike the AR-LLM,
the MDLM can condition a masked position on observed tokens on either side of
that position. However, applying the independent corruption in
Equation~\eqref{eq:mdlm-forward-process} directly to flattened hierarchical
SIDs can expose a fine SID component while masking an earlier, coarser
component. Section~\ref{sec:methodology} introduces HPD, whose structured
corruption process removes this inconsistency while retaining bidirectional
context between SIDs. We use \emph{dLLM} for a large pretrained model instantiated with
an MDLM-style objective; LLaDA and Dream are the two dLLM backbones in our
experiments~\citep{nie2025llada,ye2025dream}.

\section{Hierarchy-Preserving Diffusion}
\label{sec:methodology}

A direct application of masked discrete diffusion treats every token in a
flattened SID sequence as an independently corruptible coordinate. This
treatment is appropriate across items, where bidirectional context can expose
global relations among a user's interactions, but it ignores the coarse-to-fine
order within each hierarchical SID. HPD therefore uses a masking process with
two levels of dependency: bidirectional conditioning between SIDs and
sequential conditioning among the components of each SID.

\subsection{Two-Level Dependency Structure}
\label{sec:two-level-dependency}

Recall that $\mathbf{s}_j=(s_j^{(1)},\ldots,s_j^{(m)})$ denotes the SID of
the $j$-th item in a training sequence, where the component index
$r\in\{1,\ldots,m\}$ follows the coarse-to-fine order imposed by residual
quantization. The flattened clean sequence $\mathbf{x}$ contains $n+1$ such
SIDs: $n$ historical items and one target item.

\paragraph{Inter-SID bidirectional dependency.}
The model uses full self-attention when predicting a masked component. Thus, an
observed component of $\mathbf{s}_j$ can condition the reconstruction of a
component of $\mathbf{s}_{j'}$ for any distinct item indices $j$ and $j'$,
regardless of their relative positions in $\mathbf{x}$. In contrast to an
autoregressive model, the resulting dependency structure does not impose a
global left-to-right order across items. This bidirectional context is intended
to capture relations among SIDs throughout the interaction sequence.

\paragraph{Intra-SID sequential dependency.}
Within a single SID, later components refine the residual information left by
earlier components. Consequently, predicting $s_j^{(r)}$ while exposing a
finer component $s_j^{(r')}$ with $r'>r$ would reverse the order used to
construct the SID. We prevent this information leakage by requiring the masked
components of every SID to form a suffix. Equivalently, the observed components
of a partially corrupted SID must form a prefix.

\subsection{Prefix-Independent Masking}
\label{sec:prefix-independent-masking}

We next formalize the corruption process that realizes these two dependency
levels. For each item index $j\in\{1,\ldots,n+1\}$ and hierarchy level
$r\in\{1,\ldots,m\}$, let $\widetilde b_{j,r}\in\{0,1\}$ be a raw mask
indicator sampled independently as
$\widetilde b_{j,r}\sim\operatorname{Bernoulli}(t)$, where
$t\sim\mathcal{U}[0,1]$ is the sampled masking rate. We obtain the filtered mask
indicator $b_{j,r}$ by taking the left-to-right cumulative maximum within each
SID:
\begin{equation}
    b_{j,r}
    = \max_{1\leq q\leq r}\widetilde b_{j,q}.
    \label{eq:prefix-independent-filter}
\end{equation}
The corresponding corrupted SID component is
\begin{equation}
    z_{t,j}^{(r)}
    =
    \begin{cases}
        \mathrm{M}, & b_{j,r}=1,\\
        s_j^{(r)}, & b_{j,r}=0,
    \end{cases}
    \label{eq:prefix-independent-corruption}
\end{equation}
where $\mathrm{M}$ is the mask token. Equations~\eqref{eq:prefix-independent-filter}
and~\eqref{eq:prefix-independent-corruption} induce the prefix-independent
corruption distribution
$q_t^{\mathrm{PI}}(\mathbf{z}_t\mid\mathbf{x})$ over the flattened corrupted
sequence $\mathbf{z}_t$. The filtering is applied independently to each SID;
it does not introduce a causal order between different SIDs.

More precisely, a mask pattern $\mathbf{b}_j=(b_{j,1},\ldots,b_{j,m})$ is
valid if and only if
\begin{equation}
    b_{j,1}\leq b_{j,2}\leq\cdots\leq b_{j,m}.
    \label{eq:valid-sid-mask}
\end{equation}
For a three-component SID, the valid patterns are therefore
$(0,0,0)$, $(0,0,1)$, $(0,1,1)$, and $(1,1,1)$, where one denotes a masked
component. For example, the raw corruption
$(s_j^{(1)},\mathrm{M},s_j^{(3)})$ is converted to
$(s_j^{(1)},\mathrm{M},\mathrm{M})$. The filter preserves a visible hierarchical
prefix while allowing arbitrary SIDs elsewhere in the sequence to provide
bidirectional context.

We train the model on the filtered corruptions by replacing the independent
corruption distribution in the standard MDLM objective with
$q_t^{\mathrm{PI}}$:
\begin{equation}
    \mathcal{L}_{\mathrm{PI}}(\theta)
    = -\mathbb{E}_{\substack{(Q,Y)\sim\mathcal{D},\;t\sim\mathcal{U}[0,1],\\
        \mathbf{z}_t\sim q_t^{\mathrm{PI}}(\cdot\mid\mathbf{x})}}
    \left[
        \frac{1}{t}\sum_{k=1}^{L}
        \mathbb{I}[z_t^k=\mathrm{M}]
        \log \pi_\theta(x^k\mid\mathbf{z}_t)
    \right].
    \label{eq:prefix-independent-objective}
\end{equation}
Here, $\theta$ denotes the model parameters, $z_t^k$ and $x^k$ are the
$k$-th components of $\mathbf{z}_t$ and $\mathbf{x}$, respectively, and all
remaining symbols are defined in Section~\ref{sec:preliminaries}. The loss is
computed only at masked coordinates. Because the network attends to the full
corrupted sequence, Equation~\eqref{eq:prefix-independent-objective} retains
bidirectional learning across SIDs; because the filter masks every component
after the first masked component within an SID, it preserves the SID's
coarse-to-fine order.

\subsection{HPD Training}
\label{sec:hpd-training}

HPD must both place the newly introduced SID tokens in the dLLM's
representation space and train the backbone on recommendation sequences. Let
$\theta=(\theta_{\mathrm{SID}},\theta_{\mathrm{base}})$ partition the model
parameters. The subset $\theta_{\mathrm{SID}}$ contains the input-embedding
rows of the new SID tokens and, when the output projection is untied, their
corresponding output rows. The subset $\theta_{\mathrm{base}}$ contains all
remaining backbone parameters.

The proposed pipeline first optimizes only $\theta_{\mathrm{SID}}$ while
$\theta_{\mathrm{base}}$ remains fixed. This restriction trains the new
SID-token embeddings and output weights without changing the backbone. The
pipeline then unfreezes $\theta_{\mathrm{base}}$ and jointly optimizes all
parameters $\theta$, allowing the model to learn inter-SID relations and the
coarse-to-fine structure within each SID. The corruption process and
recommendation data remain unchanged as the set of trainable parameters
expands. The ablation in Section~\ref{sec:ablation-study} compares the proposed
pipeline with its SID-token-only variant.

\subsection{HPD Inference}
\label{sec:hpd-inference}

At inference time, HPD recovers the SID of the next item while
preserving the same two-level dependency structure used during training. We
therefore expose the complete interaction history $Q$ and initialize the target
SID as a sequence of $m$ mask tokens:
\begin{equation}
    Y^{(0)}=(\mathrm{M},\ldots,\mathrm{M}),
    \qquad
    \mathbf{z}^{(0)}=\operatorname{flat}\!\left(Q\oplus Y^{(0)}\right).
    \label{eq:inference-initialization}
\end{equation}
Here, $Y^{(0)}$ is the initial target state and $\mathbf{z}^{(0)}$ is the
corresponding model input. The observed history remains fixed throughout
decoding. The model then unmasks the target components from coarse to fine, so
every intermediate target contains an observed prefix followed by a masked
suffix.

To ensure that decoding produces only catalog items, we constrain every
expansion with the set of valid item SIDs. Let
$\mathcal{S}=\{\operatorname{SID}(i):i\in\mathcal{I}\}$ denote the catalog SID
set.
For a catalog SID $\mathbf{s}=(s^{(1)},\ldots,s^{(m)})\in\mathcal{S}$, let
$\mathbf{s}^{<r}=(s^{(1)},\ldots,s^{(r-1)})$ denote its first $r-1$
components. At each hierarchy level $r$, the catalog determines the set of
length-$(r-1)$ prefixes and the components that follow each prefix:
\begin{equation}
    \begin{aligned}
        \mathcal{P}_{r-1}
        &=\left\{\mathbf{s}^{<r}:\mathbf{s}\in\mathcal{S}\right\},\\
        \mathcal{A}_r(\mathbf{p})
        &=\left\{
            s^{(r)}:
            \substack{\mathbf{s}\in\mathcal{S},\\
                      \mathbf{s}^{<r}=\mathbf{p}}
          \right\},
        \quad \mathbf{p}\in\mathcal{P}_{r-1}.
    \end{aligned}
    \label{eq:allowed-sid-tokens}
\end{equation}
Thus, $\mathcal{P}_{r-1}$ contains \textit{every} prefix that occurs in the catalog,
and $\mathcal{A}_r(\mathbf{p})$ contains the SID components that may occur
at level $r$ after prefix $\mathbf{p}$. Both quantities are obtained from the
known catalog SIDs; they do not assume which prefix belongs to the unknown
target.

Decoding begins at level $r=1$ with an empty prefix and a fully masked target.
At hierarchy level $r\in\{1,\ldots,m\}$, beam $b$ obtains its current prefix
from the tokens selected along that beam during the preceding $r-1$ decoding
iterations. We denote this beam-specific decoded prefix by
$\widehat{\mathbf{s}}_b^{<r}=(\widehat{s}_b^{(1)},\ldots,
\widehat{s}_b^{(r-1)})$ and form the model input
\begin{equation}
    \mathbf{z}_b^{(r-1)}
    =\operatorname{flat}\!\left(
        Q\oplus
        (\widehat{s}_b^{(1)},\ldots,\widehat{s}_b^{(r-1)},
        \underbrace{\mathrm{M},\ldots,\mathrm{M}}_{m-r+1})
      \right).
    \label{eq:inference-state}
\end{equation}
The beam queries the catalog-defined set
$\mathcal{A}_r(\widehat{\mathbf{s}}_b^{<r})$ for its allowed tokens at level
$r$. At $r=1$, $\widehat{\mathbf{s}}_b^{<1}=()$ and
$\mathcal{P}_0=\{()\}$, so $\mathcal{A}_1(())$ contains every valid first-level
component. If the beam selects
$a\in\mathcal{A}_r(\widehat{\mathbf{s}}_b^{<r})$, its prefix for the next
iteration is
\begin{equation}
    \widehat{\mathbf{s}}_{b,a}^{<r+1}
    =\widehat{\mathbf{s}}_b^{<r}\oplus a.
    \label{eq:beam-prefix-update}
\end{equation}
Because $a$ is drawn from $\mathcal{A}_r(\widehat{\mathbf{s}}_b^{<r})$, the
expanded prefix belongs to $\mathcal{P}_r$ and therefore matches at least one
catalog SID. The decoder repeats this construction until all $m$ components
have been selected.

We rank the constrained hypotheses with left-to-right beam search. Let
$c_b^{(r-1)}\in\mathbb{R}$ be the cumulative log-score of beam $b$ before level
$r$, initialized as $c_1^{(0)}=0$. For each allowed token
$a\in\mathcal{A}_r(\widehat{\mathbf{s}}_b^{<r})$, its expanded score is
\begin{equation}
    c_{b,a}^{(r)}
    =c_b^{(r-1)}
     +\log\pi_\theta\!\left(
        a\mid\mathbf{z}_b^{(r-1)},\,k_r
      \right),
    \qquad k_r=mn+r,
    \label{eq:inference-beam-score}
\end{equation}
where $k_r$ is the flattened position of the $r$-th target component and
$\pi_\theta(a\mid\mathbf{z},k)$ denotes the model probability assigned to token
$a$ at position $k$ given input $\mathbf{z}$. Each beam is expanded with at
most $B$ highest-scoring allowed tokens, where $B\in\mathbb{N}$ is the branch
factor. The expanded hypotheses are pooled, and only the $W\in\mathbb{N}$
highest-scoring hypotheses are retained, where $W$ is the beam width. After
$m$ iterations, the decoder returns up to $W$ complete SIDs ranked by their
cumulative scores.

This procedure connects the training constraint to generation. Immediately
before decoding level $r$, the target mask pattern is
$(0^{r-1},1^{m-r+1})$, which satisfies Equation~\eqref{eq:valid-sid-mask}.
Consequently, no finer target component is exposed when predicting a coarser
one. At the same time, full self-attention allows every target prediction to
condition on all observed SIDs in $Q$. The decoder therefore retains sequential
coarse-to-fine generation within the target SID while exploiting bidirectional
relations across the interaction history.

\paragraph{End-to-end procedure.}
Algorithm~\ref{alg:end-to-end-method} summarizes SID construction, model
training, and inference for one test history. Let
$\mathcal{D}_{\mathrm{item}}$ denote raw training examples expressed as item
histories and next-item targets. Let
$T_{\mathrm{SID}},T_{\mathrm{all}}\in\mathbb{N}$ be the numbers of updates
before and after all model parameters become trainable, respectively, and let
$H_\star=(i_1^\star,\ldots,i_n^\star)$ be a test interaction history. The
algorithm converts $\mathcal{D}_{\mathrm{item}}$ into the empirical SID-pair
distribution $\mathcal{D}$ defined in Section~\ref{sec:problem-formulation}.
For concision, $\operatorname{Top}_K$ denotes selection of the $K$ candidates
with the largest scores.

\begin{algorithm}[t]
\caption{End-to-end Hierarchy-Preserving Diffusion (HPD) for Generative Recommendation}
\label{alg:end-to-end-method}
\small
\begin{algorithmic}[1]
\Require Item catalog $\mathcal{I}$; descriptions $\{d_i\}_{i\in\mathcal{I}}$;
raw interactions $\mathcal{D}_{\mathrm{item}}$; base dLLM $\pi_\theta$;
depth $m$; step budgets $T_{\mathrm{SID}},T_{\mathrm{all}}$; branch factor
$B$; beam width $W$; test history $H_\star$
\Ensure Trained parameters $\theta$ and ranked next-item SIDs $\mathcal{B}_m$
\Statex \textbf{SID construction}
\For{each $i\in\mathcal{I}$}
    \State $\mathbf{h}_i\gets g_{\mathrm{enc}}(d_i)$
\EndFor
\State Apply RQ-KMeans to $\{\mathbf{h}_i\}_{i\in\mathcal{I}}$ using
Equations~\eqref{eq:rq-assignment}--\eqref{eq:rq-residual}
\State Obtain $\mathbf{s}_i=\operatorname{SID}(i)$ for every item and extend the
model vocabulary with level-specific SID tokens
\State Let $\mathcal{D}$ be the empirical distribution obtained by replacing
each item in $\mathcal{D}_{\mathrm{item}}$ with its SID; set
$\mathcal{S}\gets\{\mathbf{s}_i:i\in\mathcal{I}\}$
\State Build the catalog prefix constraints $\mathcal{A}_r$ using
Equation~\eqref{eq:allowed-sid-tokens}
\Statex \textbf{HPD training}
\State Optimize only $\theta_{\mathrm{SID}}$ for $T_{\mathrm{SID}}$ updates
using stochastic gradients of $\mathcal{L}_{\mathrm{PI}}$, while holding
$\theta_{\mathrm{base}}$ fixed
\State Optimize all parameters $\theta$ for $T_{\mathrm{all}}$ updates using
the same objective
\Statex \textbf{Catalog-constrained inference}
\State $Q_\star\gets[\operatorname{SID}(i_1^\star),\ldots,
\operatorname{SID}(i_n^\star)]$ and $\mathcal{B}_0\gets\{((),0)\}$
\For{$r=1,\ldots,m$}
    \State $\mathcal{E}_r\gets\varnothing$
    \For{each $(\widehat{\mathbf{s}}^{<r},c)\in\mathcal{B}_{r-1}$}
        \State Construct $\mathbf{z}^{(r-1)}$ from $Q_\star$ and
        $\widehat{\mathbf{s}}^{<r}$ using Equation~\eqref{eq:inference-state}
        \State $\mathcal{T}\gets\operatorname{Top}_B\{(a,
        \log\pi_\theta(a\mid\mathbf{z}^{(r-1)},k_r)):
        a\in\mathcal{A}_r(\widehat{\mathbf{s}}^{<r})\}$
        \For{each $(a,\delta)\in\mathcal{T}$}
            \State Add $(\widehat{\mathbf{s}}^{<r}\oplus a,c+\delta)$ to
            $\mathcal{E}_r$
        \EndFor
    \EndFor
    \State $\mathcal{B}_r\gets\operatorname{Top}_W(\mathcal{E}_r)$
\EndFor
\State \Return $\theta$ and $\mathcal{B}_m$
\end{algorithmic}
\end{algorithm}

\section{Experiments}
\label{sec:experiments}

We evaluate two questions: how HPD compares at the final checkpoint with causal AR-LLM and standard MDLM across domains and dLLM backbones (LLaDA and Dream), as well as with
DiffGRM for Dream; and how the proposed HPD pipeline compares with its
SID-token-only variant.
After defining the datasets, model pairs, metrics, and shared protocol, we
present the final-checkpoint comparison in Section~\ref{sec:main-results} and
the component ablation in Section~\ref{sec:ablation-study}.
Overall, the proposed HPD pipeline improves over the matched autoregressive
baseline across both backbone pairs, obtains the best or tied-best result
throughout the 8B LLaDA evaluation, and outperforms the SID-token-only variant
across every metric for Dream-7B-Base on Industrial and LLaDA-8B-Base on Baby.

\subsection{Experimental Setup}
\label{sec:experimental-setup}

We evaluate whether HPD provides a consistent advantage for next-item
recommendation when applied to dLLM backbones across domains, backbone
families, and model scales. The comparison covers AR-LLMs trained with causal
next-token prediction and dLLM backbones trained with HPD's prefix-independent
MDLM objective or alternative masked discrete diffusion objectives under
otherwise matched settings.

\paragraph{Datasets.}
We conduct experiments on the Amazon Reviews 2023 collection
\citep{hou2024amazon}\footnote{\url{https://amazon-reviews-2023.github.io/}}, using the
\textit{Industrial and Scientific}, \textit{Office Products}, and
\textit{Baby Products} domains. 

\begin{wraptable}{r}{0.5\textwidth}
\caption{Statistics of the processed Amazon Reviews 2023 domains. Split columns
report numbers of next-item examples.}
\label{tab:dataset-statistics}
\centering
\small
\setlength{\tabcolsep}{3.8pt}
\resizebox{\linewidth}{!}{%
\begin{tabular}{@{}lrrrr@{}}
\toprule
\textbf{Domain} & \textbf{Train} & \textbf{Valid.} & \textbf{Test} & \textbf{Items} \\
\midrule
Industrial and Scientific & 45,123 & 5,640 & 5,641 & 4,980 \\
Office Products            & 402,257 & 50,282 & 50,283 & 25,862 \\
Baby Products              & 396,677 & 49,585 & 49,585 & 16,947 \\
\bottomrule
\end{tabular}%
}
\end{wraptable}

For each domain, we use processed 5-core slices
spanning October 2016 through November 2018, where the
5-core filter removes users and items with fewer than five interactions. Each
processed row forms one next-item example consisting of an interaction history
and its subsequent target. The examples are divided approximately $80/10/10$
into training, validation, and test sets. Table~\ref{tab:dataset-statistics}
reports the resulting split and catalog sizes. We abbreviate the three domains
as Industrial, Office, and Baby in the remaining discussion.

\paragraph{Base models and comparison methods.}
We study two approximately size-matched AR-LLM/dLLM pairs. The 8B-parameter
comparison pairs the Qwen3-8B AR-LLM~\citep{yang2025qwen3} with the
LLaDA-8B-Base dLLM~\citep{nie2025llada}; the 7B-parameter comparison pairs the
Qwen2.5-7B AR-LLM~\citep{yang2024qwen25} with the Dream-7B-Base
dLLM~\citep{ye2025dream}.
Each backbone uses its corresponding three-level L3-K1024 SID vocabulary. For
the base-model controls, we run inference immediately after adding the new SID
tokens to the embedding and output spaces, before recommendation training. We
then compare the following recommendation-training objectives:
\begin{itemize}
    \item \textbf{AR-LLM} trains Qwen3 and Qwen2.5 with the causal objective in
    Equation~\eqref{eq:ar-objective}, following OpenOneRec, and provides the
    autoregressive baseline~\citep{zhou2025openonerec}.
    \item \textbf{HPD} trains the LLaDA and Dream dLLMs with the
    prefix-independent corruption and objective in
    Equation~\eqref{eq:prefix-independent-objective}.
    \item \textbf{MDLM} trains the same dLLM backbones with independent token
    masking and the standard MDLM objective in
    Equations~\eqref{eq:mdlm-forward-process}
    and~\eqref{eq:mdlm-objective}~\citep{nie2025llada}. This comparison
    directly contrasts HPD's intra-SID prefix constraint with independent
    token masking.
    \item \textbf{DiffGRM-OCN} replaces the masking rule with the on-policy
    coherent noising objective of DiffGRM~\citep{liu2026diffgrm}, which focuses
    supervision on low-confidence SID components. We include this objective for
    the Dream backbone and denote it as \emph{DiffGRM} in the result tables.
\end{itemize}
All methods in the main comparison use the proposed training pipeline described
in Section~\ref{sec:hpd-training}. The pipeline first optimizes only the new
SID-token embeddings and output weights and then trains all model parameters.
Main results report the final checkpoint after the proposed pipeline concludes.
The ablation configuration is defined in Section~\ref{sec:ablation-study}.

\paragraph{Evaluation metrics.}
We measure top-$K$ next-item accuracy with Hit Rate (HR@$K$) and Normalized
Discounted Cumulative Gain (NDCG@$K$) for $K\in\{3,5,10\}$. Let
$\mathcal{T}$ be the test set and let
$\rho_u\in\mathbb{N}\cup\{\infty\}$ be the one-indexed rank of the ground-truth
target for test example $u\in\mathcal{T}$, with
$\rho_u=\infty$ when the target is absent from the returned list. Because each
example has one relevant target, the metrics reduce to
\begin{equation}
    \operatorname{HR@}K
    =\frac{1}{|\mathcal{T}|}\sum_{u\in\mathcal{T}}
      \mathbb{I}[\rho_u\leq K],
    \qquad
    \operatorname{NDCG@}K
    =\frac{1}{|\mathcal{T}|}\sum_{u\in\mathcal{T}}
      \frac{\mathbb{I}[\rho_u\leq K]}{\log_2(\rho_u+1)}.
    \label{eq:evaluation-metrics}
\end{equation}
HR@$K$ records whether the target appears among the first $K$ predictions,
whereas NDCG@$K$ additionally rewards placing a hit closer to the top of the
ranked list.

\paragraph{Implementation details.}
We train every model on eight NVIDIA H200 GPUs with fully sharded data
parallelism, bfloat16 mixed precision, and TF32 matrix operations. Optimization
uses AdamW, a cosine learning-rate schedule, and gradient clipping at norm
$1.0$. For the first 1,000 updates, only the new SID-token embeddings and
output weights are trainable; the learning rate is $2\times10^{-4}$, weight
decay is $0.01$, and 20 updates are used for warmup. For the next 1,000
updates, all model parameters are trainable, with a learning rate of
$1\times10^{-5}$, weight decay of $0.1$, and 100 warmup updates. We evaluate
the final checkpoint after training concludes.

Each GPU processes one packed sequence at a time. Accumulating gradients over
two steps gives the Qwen runs an effective batch of 16 packed sequences across
eight GPUs. While optimizing only the SID-token parameters, both Dream and
LLaDA accumulate gradients over four steps, yielding an effective batch of 32.
During training of all model parameters, Dream retains this four-step
accumulation schedule, whereas LLaDA uses two steps and therefore an effective
batch of 16. We cap
packed training sequences at 32,768 tokens for Qwen and 8,192 tokens for Dream.
For LLaDA, the cap is 4,096 tokens while optimizing only the SID-token
parameters and 2,048 tokens while training all model parameters. Training and
data shuffling use seed 1024.
For runs with the DiffGRM-OCN objective, we use maximum softmax probability to
identify the least-confident components, determine the number of guided steps
automatically, and apply label smoothing of $0.1$. We do not refresh the
confidence estimates after every guided step, and set the OCN codebook-size
parameter to 1,025 to match the instantiated SID-token range. All other
optimization settings are shared with the corresponding dLLM run.

At evaluation time, we retain at most the most recent 2,560 prompt tokens and
constrain every generated SID to the item catalog. Autoregressive models decode
with 50 beams, a per-device evaluation batch size of eight, and a limit of 32
new tokens. The HPD-trained dLLMs use the procedure in
Section~\ref{sec:hpd-inference}: they unmask the three target
components from coarse to fine, expand each hypothesis with at most 50 valid
tokens, and retain the 50 highest-scoring beams after every level. Thus, both
model families return up to 50 catalog-valid candidates and are evaluated with
the same ranking protocol.

\subsection{Main Experimental Results}
\label{sec:main-results}

Tables~\ref{tab:qwen_llada_final_results}
and~\ref{tab:qwen_dream_final_results} report recommendation quality at the
final checkpoint after training concludes. The parenthesized values quantify
the change from the corresponding base-model row. \textbf{Bold} and \underline{underlined} values
denote the best and second-best distinct trained results within each domain,
respectively; ties receive the same emphasis. The analysis below compares the
final metric values directly.

\paragraph{Base-model reference.}
The base-model rows yield metrics close to zero across all three domains.
These checkpoints are evaluated immediately after the SID tokens are added to
the vocabulary and before recommendation training. The vocabulary-expansion
routine randomly samples the new input embeddings from a multivariate normal
distribution matched to the pretrained embeddings' mean and covariance, and
samples the untied output rows from a normal distribution matched to the
pretrained output rows' elementwise mean and standard deviation. Because these
new parameters have not yet been aligned with the backbone's representation
space or trained to capture catalog semantics, their performance is consistent
with effectively uninformed, near-random prediction. We treat these rows as an
initialization reference rather than a trained recommendation baseline.

\begin{table}[!htb]
\centering
\caption{Final recommendation results for the Qwen3/LLaDA model pair, grouped by backbone.}
\label{tab:qwen_llada_final_results}
\small
\setlength{\tabcolsep}{3.5pt}
\renewcommand{\arraystretch}{1.15}
\resizebox{0.8\linewidth}{!}{%
\begin{tabular}{lcccccc}
\toprule
\textbf{Method}
& \textbf{HR@3} & \textbf{HR@5} & \textbf{HR@10}
& \textbf{NDCG@3} & \textbf{NDCG@5} & \textbf{NDCG@10} \\
\midrule
\multicolumn{7}{@{}l}{\textbf{Industrial and Scientific}} \\
\addlinespace[2pt]
\textit{Qwen3-8B}
& 0.0000 & 0.0004 & 0.0012 & 0.0000 & 0.0002 & 0.0004 \\
\addlinespace[1pt]
\multirow{2}{*}{\hspace*{0.75em}AR-LLM (Qwen3-8B)}
& 0.0092 & 0.0129 & 0.0261 & 0.0071 & 0.0086 & 0.0128 \\
& {\scriptsize (0.0092)}
& {\scriptsize (0.0125)}
& {\scriptsize (0.0249)}
& {\scriptsize (0.0071)}
& {\scriptsize (0.0084)}
& {\scriptsize (0.0124)} \\
\addlinespace[2pt]
\textit{LLaDA-8B-Base}
& 0.0005 & 0.0009 & 0.0021 & 0.0005 & 0.0006 & 0.0010 \\
\addlinespace[1pt]
\multirow{2}{*}{\hspace*{0.75em}MDLM (LLaDA-8B-Base)}
& \underline{0.0161}
& \underline{0.0207}
& \underline{0.0289}
& \underline{0.0123}
& \underline{0.0142}
& \underline{0.0168} \\
& {\scriptsize\underline{(0.0156)}}
& {\scriptsize\underline{(0.0198)}}
& {\scriptsize\underline{(0.0268)}}
& {\scriptsize\underline{(0.0118)}}
& {\scriptsize\underline{(0.0136)}}
& {\scriptsize\underline{(0.0158)}} \\
\rowcolor[HTML]{F0F8FF}
& \textbf{0.0266}
& \textbf{0.0309}
& \textbf{0.0374}
& \textbf{0.0239}
& \textbf{0.0256}
& \textbf{0.0277} \\
\rowcolor[HTML]{F0F8FF}
\multirow{-2}{*}{\hspace*{0.75em}HPD (LLaDA-8B-Base)}
& {\scriptsize\textbf{(0.0261)}}
& {\scriptsize\textbf{(0.0300)}}
& {\scriptsize\textbf{(0.0353)}}
& {\scriptsize\textbf{(0.0234)}}
& {\scriptsize\textbf{(0.0250)}}
& {\scriptsize\textbf{(0.0267)}} \\
\midrule
\multicolumn{7}{@{}l}{\textbf{Office Products}} \\
\addlinespace[2pt]
\textit{Qwen3-8B}
& 0.0002 & 0.0004 & 0.0008 & 0.0002 & 0.0002 & 0.0004 \\
\addlinespace[1pt]
\multirow{2}{*}{\hspace*{0.75em}AR-LLM (Qwen3-8B)}
& 0.0023 & 0.0036 & 0.0065 & 0.0016 & 0.0022 & 0.0031 \\
& {\scriptsize (0.0021)}
& {\scriptsize (0.0032)}
& {\scriptsize (0.0057)}
& {\scriptsize (0.0014)}
& {\scriptsize (0.0020)}
& {\scriptsize (0.0027)} \\
\addlinespace[2pt]
\textit{LLaDA-8B-Base}
& 0.0001 & 0.0002 & 0.0004 & 0.0001 & 0.0001 & 0.0002 \\
\addlinespace[1pt]
\multirow{2}{*}{\hspace*{0.75em}MDLM (LLaDA-8B-Base)}
& \underline{0.0053}
& \underline{0.0075}
& \underline{0.0124}
& \underline{0.0040}
& \underline{0.0049}
& \underline{0.0064} \\
& {\scriptsize\underline{(0.0052)}}
& {\scriptsize\underline{(0.0073)}}
& {\scriptsize\underline{(0.0120)}}
& {\scriptsize\underline{(0.0039)}}
& {\scriptsize\underline{(0.0048)}}
& {\scriptsize\underline{(0.0062)}} \\
\rowcolor[HTML]{F0F8FF}
& \textbf{0.0060}
& \textbf{0.0085}
& \textbf{0.0134}
& \textbf{0.0047}
& \textbf{0.0057}
& \textbf{0.0073} \\
\rowcolor[HTML]{F0F8FF}
\multirow{-2}{*}{\hspace*{0.75em}HPD (LLaDA-8B-Base)}
& {\scriptsize\textbf{(0.0059)}}
& {\scriptsize\textbf{(0.0083)}}
& {\scriptsize\textbf{(0.0130)}}
& {\scriptsize\textbf{(0.0046)}}
& {\scriptsize\textbf{(0.0056)}}
& {\scriptsize\textbf{(0.0071)}} \\
\midrule
\multicolumn{7}{@{}l}{\textbf{Baby Products}} \\
\addlinespace[2pt]
\textit{Qwen3-8B}
& 0.0003 & 0.0004 & 0.0007 & 0.0002 & 0.0003 & 0.0003 \\
\addlinespace[1pt]
\multirow{2}{*}{\hspace*{0.75em}AR-LLM (Qwen3-8B)}
& 0.0021
& 0.0038
& 0.0069
& \underline{0.0014}
& 0.0022
& 0.0032 \\
& {\scriptsize (0.0018)}
& {\scriptsize (0.0034)}
& {\scriptsize (0.0062)}
& {\scriptsize\underline{(0.0012)}}
& {\scriptsize (0.0019)}
& {\scriptsize (0.0029)} \\
\addlinespace[2pt]
\textit{LLaDA-8B-Base}
& 0.0006 & 0.0009 & 0.0013 & 0.0004 & 0.0006 & 0.0007 \\
\addlinespace[1pt]
\multirow{2}{*}{\hspace*{0.75em}MDLM (LLaDA-8B-Base)}
& \underline{0.0092}
& \underline{0.0125}
& \underline{0.0202}
& \textbf{0.0074}
& \underline{0.0087}
& \underline{0.0112} \\
& {\scriptsize\underline{(0.0086)}}
& {\scriptsize\underline{(0.0116)}}
& {\scriptsize\underline{(0.0189)}}
& {\scriptsize\textbf{(0.0070)}}
& {\scriptsize\underline{(0.0081)}}
& {\scriptsize\underline{(0.0105)}} \\
\rowcolor[HTML]{F0F8FF}
& \textbf{0.0095}
& \textbf{0.0133}
& \textbf{0.0207}
& \textbf{0.0074}
& \textbf{0.0090}
& \textbf{0.0114} \\
\rowcolor[HTML]{F0F8FF}
\multirow{-2}{*}{\hspace*{0.75em}HPD (LLaDA-8B-Base)}
& {\scriptsize\textbf{(0.0089)}}
& {\scriptsize\textbf{(0.0124)}}
& {\scriptsize\textbf{(0.0194)}}
& {\scriptsize\textbf{(0.0070)}}
& {\scriptsize\textbf{(0.0084)}}
& {\scriptsize\textbf{(0.0107)}} \\
\bottomrule
\end{tabular}%
}
\end{table}

\paragraph{Results with the 8B backbones.}
HPD obtains the best or tied-best score for all 18 metric--domain combinations
in the LLaDA-8B-Base comparison. On Industrial, HPD improves HR@10 from
$0.0261$ for AR-LLM and $0.0289$ for standard MDLM to $0.0374$; its NDCG@10
reaches $0.0277$, compared with $0.0128$ and $0.0168$, respectively. The same
ordering holds on Office, where HPD attains $0.0134$ HR@10 and $0.0073$
NDCG@10. On Baby, HPD and MDLM are closer, but HPD achieves the best
performance at five cutoffs and ties MDLM at $0.0074$ NDCG@3. The consistent
advantage over the matched AR-LLM indicates that bidirectional masked discrete diffusion is
effective for these SID sequences. The additional gains over MDLM, especially
on Industrial, are consistent with the intra-SID prefix constraint providing
useful structure for the 8B dLLM.

\begin{table}[!htb]
\centering
\caption{Final recommendation results for the Qwen2.5/Dream model pair, grouped by backbone.}
\label{tab:qwen_dream_final_results}
\small
\setlength{\tabcolsep}{3.5pt}
\renewcommand{\arraystretch}{1.15}
\resizebox{0.8\linewidth}{!}{%
\begin{tabular}{lcccccc}
\toprule
\textbf{Method}
& \textbf{HR@3} & \textbf{HR@5} & \textbf{HR@10}
& \textbf{NDCG@3} & \textbf{NDCG@5} & \textbf{NDCG@10} \\
\midrule
\multicolumn{7}{@{}l}{\textbf{Industrial and Scientific}} \\
\addlinespace[2pt]
\textit{Qwen2.5-7B}
& 0.0005 & 0.0009 & 0.0025 & 0.0004 & 0.0006 & 0.0011 \\
\addlinespace[1pt]
\multirow{2}{*}{\hspace*{0.75em}AR-LLM (Qwen2.5-7B)}
& 0.0184 & 0.0236 & 0.0285 & 0.0152 & 0.0173 & 0.0189 \\
& {\scriptsize (0.0179)}
& {\scriptsize (0.0227)}
& {\scriptsize (0.0260)}
& {\scriptsize (0.0148)}
& {\scriptsize (0.0167)}
& {\scriptsize (0.0178)} \\
\addlinespace[2pt]
\textit{Dream-7B-Base}
& 0.0000 & 0.0004 & 0.0012 & 0.0000 & 0.0001 & 0.0004 \\
\addlinespace[1pt]
\multirow{2}{*}{\hspace*{0.75em}MDLM (Dream-7B-Base)}
& \textbf{0.0516}
& \underline{0.0633}
& \underline{0.0846}
& \textbf{0.0421}
& \underline{0.0468}
& \underline{0.0536} \\
& {\scriptsize\textbf{(0.0516)}}
& {\scriptsize\underline{(0.0629)}}
& {\scriptsize\underline{(0.0834)}}
& {\scriptsize\textbf{(0.0421)}}
& {\scriptsize\underline{(0.0467)}}
& {\scriptsize\underline{(0.0532)}} \\
\multirow{2}{*}{\hspace*{0.75em}DiffGRM (Dream-7B-Base)}
& 0.0434 & 0.0507 & 0.0605 & 0.0371 & 0.0401 & 0.0433 \\
& {\scriptsize (0.0434)}
& {\scriptsize (0.0503)}
& {\scriptsize (0.0593)}
& {\scriptsize (0.0371)}
& {\scriptsize (0.0400)}
& {\scriptsize (0.0429)} \\
\rowcolor[HTML]{F0F8FF}
& \underline{0.0505}
& \textbf{0.0644}
& \textbf{0.0906}
& \underline{0.0417}
& \textbf{0.0474}
& \textbf{0.0558} \\
\rowcolor[HTML]{F0F8FF}
\multirow{-2}{*}{\hspace*{0.75em}HPD (Dream-7B-Base)}
& {\scriptsize\underline{(0.0505)}}
& {\scriptsize\textbf{(0.0640)}}
& {\scriptsize\textbf{(0.0894)}}
& {\scriptsize\underline{(0.0417)}}
& {\scriptsize\textbf{(0.0473)}}
& {\scriptsize\textbf{(0.0554)}} \\
\midrule
\multicolumn{7}{@{}l}{\textbf{Office Products}} \\
\addlinespace[2pt]
\textit{Qwen2.5-7B}
& 0.0001 & 0.0002 & 0.0005 & 0.0001 & 0.0001 & 0.0002 \\
\addlinespace[1pt]
\multirow{2}{*}{\hspace*{0.75em}AR-LLM (Qwen2.5-7B)}
& 0.0018 & 0.0027 & 0.0046 & 0.0014 & 0.0018 & 0.0024 \\
& {\scriptsize (0.0017)}
& {\scriptsize (0.0025)}
& {\scriptsize (0.0041)}
& {\scriptsize (0.0013)}
& {\scriptsize (0.0017)}
& {\scriptsize (0.0022)} \\
\addlinespace[2pt]
\textit{Dream-7B-Base}
& 0.0001 & 0.0002 & 0.0004 & 0.0001 & 0.0001 & 0.0002 \\
\addlinespace[1pt]
\multirow{2}{*}{\hspace*{0.75em}MDLM (Dream-7B-Base)}
& \textbf{0.0367}
& \textbf{0.0446}
& \textbf{0.0573}
& \textbf{0.0319}
& \textbf{0.0352}
& \textbf{0.0393} \\
& {\scriptsize\textbf{(0.0366)}}
& {\scriptsize\textbf{(0.0444)}}
& {\scriptsize\textbf{(0.0569)}}
& {\scriptsize\textbf{(0.0318)}}
& {\scriptsize\textbf{(0.0351)}}
& {\scriptsize\textbf{(0.0391)}} \\
\multirow{2}{*}{\hspace*{0.75em}DiffGRM (Dream-7B-Base)}
& 0.0098 & 0.0130 & 0.0194 & 0.0074 & 0.0088 & 0.0108 \\
& {\scriptsize (0.0097)}
& {\scriptsize (0.0128)}
& {\scriptsize (0.0190)}
& {\scriptsize (0.0073)}
& {\scriptsize (0.0087)}
& {\scriptsize (0.0106)} \\
\rowcolor[HTML]{F0F8FF}
& \underline{0.0334}
& \underline{0.0409}
& \underline{0.0546}
& \underline{0.0280}
& \underline{0.0310}
& \underline{0.0354} \\
\rowcolor[HTML]{F0F8FF}
\multirow{-2}{*}{\hspace*{0.75em}HPD (Dream-7B-Base)}
& {\scriptsize\underline{(0.0333)}}
& {\scriptsize\underline{(0.0407)}}
& {\scriptsize\underline{(0.0542)}}
& {\scriptsize\underline{(0.0279)}}
& {\scriptsize\underline{(0.0309)}}
& {\scriptsize\underline{(0.0352)}} \\
\midrule
\multicolumn{7}{@{}l}{\textbf{Baby Products}} \\
\addlinespace[2pt]
\textit{Qwen2.5-7B}
& 0.0003 & 0.0004 & 0.0007 & 0.0002 & 0.0002 & 0.0003 \\
\addlinespace[1pt]
\multirow{2}{*}{\hspace*{0.75em}AR-LLM (Qwen2.5-7B)}
& 0.0028 & 0.0039 & 0.0060 & 0.0022 & 0.0027 & 0.0033 \\
& {\scriptsize (0.0025)}
& {\scriptsize (0.0035)}
& {\scriptsize (0.0053)}
& {\scriptsize (0.0020)}
& {\scriptsize (0.0025)}
& {\scriptsize (0.0030)} \\
\addlinespace[2pt]
\textit{Dream-7B-Base}
& 0.0002 & 0.0003 & 0.0007 & 0.0002 & 0.0002 & 0.0003 \\
\addlinespace[1pt]
\multirow{2}{*}{\hspace*{0.75em}MDLM (Dream-7B-Base)}
& \textbf{0.0207}
& \textbf{0.0264}
& \textbf{0.0374}
& \textbf{0.0171}
& \textbf{0.0194}
& \textbf{0.0230} \\
& {\scriptsize\textbf{(0.0205)}}
& {\scriptsize\textbf{(0.0261)}}
& {\scriptsize\textbf{(0.0367)}}
& {\scriptsize\textbf{(0.0169)}}
& {\scriptsize\textbf{(0.0192)}}
& {\scriptsize\textbf{(0.0227)}} \\
\multirow{2}{*}{\hspace*{0.75em}DiffGRM (Dream-7B-Base)}
& 0.0050 & 0.0077 & 0.0125 & 0.0037 & 0.0048 & 0.0063 \\
& {\scriptsize (0.0048)}
& {\scriptsize (0.0074)}
& {\scriptsize (0.0118)}
& {\scriptsize (0.0035)}
& {\scriptsize (0.0046)}
& {\scriptsize (0.0060)} \\
\rowcolor[HTML]{F0F8FF}
& \underline{0.0178}
& \underline{0.0234}
& \underline{0.0331}
& \underline{0.0143}
& \underline{0.0167}
& \underline{0.0198} \\
\rowcolor[HTML]{F0F8FF}
\multirow{-2}{*}{\hspace*{0.75em}HPD (Dream-7B-Base)}
& {\scriptsize\underline{(0.0176)}}
& {\scriptsize\underline{(0.0231)}}
& {\scriptsize\underline{(0.0324)}}
& {\scriptsize\underline{(0.0141)}}
& {\scriptsize\underline{(0.0165)}}
& {\scriptsize\underline{(0.0195)}} \\
\bottomrule
\end{tabular}%
}
\end{table}

\paragraph{Results with the 7B backbones.}
The Dream-7B-Base results show a more domain-dependent comparison between HPD
and standard MDLM. On Industrial, HPD leads four of the six metrics, including
$0.0906$ HR@10 and $0.0558$ NDCG@10, while MDLM achieves slightly higher values
at the two top-$3$ cutoffs. On Office and Baby, MDLM achieves the best
performance across all six metrics, while HPD consistently obtains the
second-best results.
Across all three domains, HPD exceeds both AR-LLM and DiffGRM at every reported
cutoff. These results show that prefix-independent masking improves over the
causal and confidence-guided alternatives for Dream-7B-Base. Across both
backbone comparisons, HPD consistently improves over causal recommendation
training at every evaluated cutoff.


\subsection{Ablation Study}
\label{sec:ablation-study}

Tables~\ref{tab:ablation_dream_industrial}
and~\ref{tab:ablation_llada_baby} compare the proposed HPD pipeline with its
SID-token-only variant for Dream-7B-Base on Industrial and LLaDA-8B-Base on
Baby. The proposed pipeline first optimizes only the SID-token parameters and
then trains all model parameters. The SID-token-only variant omits training
with all model parameters. Bold values denote the better result in each column.

\begin{table}[!htb]
\centering
\caption{Training-pipeline ablation for HPD with Dream-7B-Base on Industrial
and Scientific.}
\label{tab:ablation_dream_industrial}
\small
\setlength{\tabcolsep}{3.5pt}
\resizebox{0.7\linewidth}{!}{%
\begin{tabular}{@{}lcccccc@{}}
\toprule
\textbf{Trainable parameters}
& \textbf{HR@3} & \textbf{HR@5} & \textbf{HR@10}
& \textbf{NDCG@3} & \textbf{NDCG@5} & \textbf{NDCG@10} \\
\midrule
SID embeddings/output weights
& 0.0466 & 0.0560 & 0.0762 & 0.0401 & 0.0440 & 0.0505 \\
All model parameters
& \textbf{0.0505} & \textbf{0.0644} & \textbf{0.0906}
& \textbf{0.0417} & \textbf{0.0474} & \textbf{0.0558} \\
\bottomrule
\end{tabular}%
}
\end{table}

\begin{table}[!htb]
\centering
\caption{Training-pipeline ablation for HPD with LLaDA-8B-Base on Baby
Products.}
\label{tab:ablation_llada_baby}
\small
\setlength{\tabcolsep}{3.5pt}
\resizebox{0.7\linewidth}{!}{%
\begin{tabular}{@{}lcccccc@{}}
\toprule
\textbf{Trainable parameters}
& \textbf{HR@3} & \textbf{HR@5} & \textbf{HR@10}
& \textbf{NDCG@3} & \textbf{NDCG@5} & \textbf{NDCG@10} \\
\midrule
SID embeddings/output weights
& 0.0031 & 0.0044 & 0.0072 & 0.0023 & 0.0028 & 0.0037 \\
All model parameters
& \textbf{0.0095} & \textbf{0.0133} & \textbf{0.0207}
& \textbf{0.0074} & \textbf{0.0090} & \textbf{0.0114} \\
\bottomrule
\end{tabular}%
}
\end{table}

The proposed HPD pipeline outperforms the SID-token-only variant on every
reported metric in both settings. On Industrial with Dream-7B-Base, HR@10
increases from $0.0762$ to $0.0906$, while NDCG@10 increases from $0.0505$ to
$0.0558$. On Baby with LLaDA-8B-Base, HR@10 increases from $0.0072$ to
$0.0207$, and NDCG@10 increases from $0.0037$ to $0.0114$. The SID-token-only
variant learns the new SID-token parameters but cannot update the backbone on
recommendation sequences. The consistent gains of the proposed pipeline show
the value of training all model parameters.

\FloatBarrier
\section{Related Works}
\label{sec:related-work}

Our work connects three lines of research: sequential and generative
recommendation, masked discrete diffusion language modeling, and masked
discrete diffusion for generative recommendation. We organize the discussion
around how each line represents items and factorizes their dependencies, since
these choices determine whether the hierarchy within a residual SID is
preserved.

\subsection{Sequential and Generative Recommendation}

Traditional sequential recommenders predict the next item directly from a
user's interaction history. GRU4Rec models the history recurrently, whereas
SASRec uses causal self-attention~\citep{hidasi2016session,kang2018self}.
BERT4Rec instead reconstructs masked items with bidirectional
self-attention~\citep{sun2019bert4rec}. Although BERT4Rec demonstrates the
value of non-causal context for recommendation, each catalog item remains one
atomic identifier; the masking process therefore does not address dependencies
among multiple codes representing the same item.

SID-based generative recommendation replaces a catalog-sized output space with
a short sequence of semantic codes. TIGER establishes this formulation with
residual-quantized item representations~\citep{rajput2023recommender}, while
OneRec and QARM extend it toward unified retrieval and ranking and multimodal
alignment, respectively~\citep{deng2025onerec,luo2025qarm}. Recent work also
studies whether the paradigm benefits from larger pretrained backbones.
MiniOneRec trains Qwen models from 0.5B to 7B parameters in an open
supervised- and reinforcement-learning pipeline, and OpenOneRec develops 1.7B-
and 8B-parameter recommendation foundation models with co-pretraining and
post-training~\citep{kong2025minionerec,zhou2025openonerec}. These systems use
an autoregressive factorization, which naturally respects the coarse-to-fine
order within a residual SID but also restricts information flow between items.
HPD retains the same hierarchical SID representation while assigning
different dependency structures to the inter-item and intra-item levels.

\subsection{Masked Discrete Diffusion Language Models}

Discrete diffusion generalizes denoising processes to categorical state
spaces~\citep{austin2021structured}. SEDD formulates discrete diffusion
training through ratios of the data distribution~\citep{lou2024discrete}, and
MDLM specializes the process to absorbing-state token masking with a simplified
likelihood-based objective~\citep{sahoo2024simple}. These methods corrupt
discrete vocabulary symbols and are therefore distinct from continuous
diffusion over token embeddings.

At larger scale, LLaDA and Dream demonstrate that masked discrete diffusion can
serve as the training basis for diffusion large language models
(dLLMs)~\citep{nie2025llada,ye2025dream}. LLaDA~1.5 introduces
variance-reduced preference optimization, while iLLaDA advances from-scratch
dLLM pretraining~\citep{zhu2025llada15,nie2026illada}. Rather than proposing a
new dLLM architecture, HPD trains the pretrained LLaDA and Dream backbones for
recommendation. Its contribution lies in matching the MDLM corruption and
decoding processes to the dependency structure of residual SIDs.

\subsection{Masked Discrete Diffusion for Generative Recommendation}

Recent methods apply masked discrete diffusion to overcome the strictly
left-to-right generation used by AR-LLM recommenders. MaskGR independently
masks SID tokens and supports flexible reconstruction orders, enabling
non-autoregressive generation without changing the SID
tokenizer~\citep{shah2025maskgr}. MindRec uses a dLLM to generate salient
preference tokens and then refines a category hierarchy and item representation
from coarse to fine~\citep{gao2025mindrec}. DiffGRM takes a different approach:
it replaces residual quantization with parallel semantic encoding to remove
sequential dependence among SID components and uses on-policy coherent noising
to focus training on uncertain components~\citep{liu2026diffgrm}.

HPD addresses a complementary setting in which the residual SID
hierarchy is retained. Prefix-independent masking makes every corrupted SID an
observed prefix followed by a masked suffix, preventing a fine residual code
from being exposed when its coarse parent is hidden. Full self-attention remains
available across items, and catalog-constrained coarse-to-fine decoding mirrors
the training dependency at inference time. This separation preserves the
existing tokenizer hierarchy rather than assuming independent SID components
or replacing them with a parallel code.

\section{Conclusion}

We introduced Hierarchy-Preserving Diffusion, a masked discrete diffusion
formulation for generative recommendation that uses bidirectional context
between item SIDs while retaining their coarse-to-fine internal dependency.
HPD enforces this structure through prefix-independent masking and mirrors it
at inference time with catalog-constrained coarse-to-fine decoding. Its
proposed training pipeline first optimizes only the newly introduced SID-token
embeddings and output weights while the backbone is fixed and then jointly
optimizes all model parameters. HPD requires no task-specific architectural
modification.

At the final checkpoint after training concludes, HPD achieves higher HR and
NDCG than its matched AR-LLM counterpart at every evaluated cutoff across two
backbone pairs and three Amazon domains. HPD also obtains the best or tied-best
score across all 18 reported metric--domain combinations in the 8B comparison.
Comparisons with standard MDLM and DiffGRM evaluate HPD against alternative
masked discrete diffusion objectives under matched settings. The proposed HPD
pipeline outperforms the SID-token-only variant on every metric for
Dream-7B-Base on Industrial and LLaDA-8B-Base on Baby. These results support
separating inter-SID and intra-SID dependencies and retaining the proposed HPD
training pipeline.

\clearpage
\newpage
\bibliographystyle{assets/plainnat}
\bibliography{sample-base}

\clearpage
\newpage


\end{document}